%% file: main.tex
\documentclass[runningheads]{llncs}
\usepackage{graphicx}
\usepackage{tikz}
\usepackage{comment}
\usepackage{amsmath,amssymb} 

\usepackage{array}
\usepackage{orcidlink}
\usepackage{color}
\newcommand{\todo}[1]{{#1}}

\usepackage[accsupp]{axessibility}  

\usepackage[absolute]{textpos}

\begin{document}

\definecolor{somegray}{rgb}{0.5, 0.5, 0.5}
\newcommand{\darkgrayed}[1]{\textcolor{somegray}{#1}}
\begin{textblock}{8}(4, 0.7)
\begin{center}
\darkgrayed{This paper has been accepted for publication at the \\
European Conference on Computer Vision (ECCV), Tel Aviv, 2022}
\end{center}
\end{textblock}

\pagestyle{headings}
\mainmatter
\def\ECCVSubNumber{7785}
\title{ESS: Learning Event-based Semantic \\
Segmentation from Still Images}

\titlerunning{ESS: Learning Event-based Semantic Segmentation from Still Images}
\author{
Zhaoning Sun\thanks{equal contribution} \orcidlink{0000-0002-9521-6562} \and
Nico Messikommer$^{\star}$ \orcidlink{0000-0003-1444-1176} \and
Daniel Gehrig \orcidlink{0000-0001-9952-3335}, \\ and
Davide Scaramuzza \orcidlink{0000-0002-3831-6778} 
}
\authorrunning{Sun et al.}
\institute{Dept. of Informatics, Univ. of Zurich and \\
Dept. of Neuroinformatics, Univ. of Zurich and ETH Zurich\\
\email{zhasun@student.ethz.ch \{nmessi,dgehrig,sdavide\}@ifi.uzh.ch}}

{\let\newpage\relax\maketitle}

\input{sections/abstract}

\input{sections/introduction}

\input{sections/related_work}
\input{sections/method}
\input{sections/results}

\input{sections/conclusion}

\input{sections/acknowledgment}

\input{sections/supplementary}
\bibliographystyle{splncs04}
\bibliography{all}
\end{document}

%% file: sections/abstract.tex
\begin{abstract}
Retrieving accurate semantic information in challenging high dynamic range (HDR) and high-speed conditions remains an open challenge for image-based algorithms due to severe image degradations.
Event cameras promise to address these challenges since they feature a much higher dynamic range and are resilient to motion blur. 
Nonetheless, semantic segmentation with event cameras is still in its infancy which is chiefly due to the lack of high-quality, labeled datasets.
In this work, we introduce \todo{ESS (Event-based Semantic Segmentation)}, which tackles this problem by directly transferring the semantic segmentation task from existing labeled image datasets to unlabeled events via unsupervised domain adaptation (UDA). 
Compared to existing UDA methods, our approach aligns recurrent, motion-invariant event embeddings with image embeddings. 
For this reason, our method neither requires video data nor per-pixel alignment between images and events and, crucially, does not need to hallucinate motion from still images. 
Additionally, we introduce DSEC-Semantic, the first large-scale event-based dataset with fine-grained labels. 
We show that using image labels alone, ESS outperforms existing UDA approaches, and when combined with event labels, it even outperforms state-of-the-art supervised approaches on both DDD17 and DSEC-Semantic. Finally, ESS is general-purpose, which unlocks the vast amount of existing labeled image datasets and paves the way for new and exciting research directions in new fields previously inaccessible for event cameras.

\keywords{Transfer learning, Low-level vision, Segmentation}
\end{abstract}

%% file: sections/introduction.tex
\noindent\textbf{Multimedia Material}
The code is available at \url{https://github.com/uzh-rpg/ess}, dataset at \url{https://dsec.ifi.uzh.ch/dsec-semantic/} and video at \url{https://youtu.be/Tby5c9IDsDc} 

\section{Introduction}

In recent years, event cameras have become attractive sensors in a wide range of applications, spanning both computer vision and robotics.
In particular, thanks to their high dynamic range, microsecond-level latency, and resilience to motion blur, algorithms leveraging event data have made various breakthroughs in fields such as Simultaneous Localization and Mapping (SLAM)~\cite{Rosinol18ral,Zhu17cvpr}, computational photography~\cite{Tulyakov21cvpr,Rebecq19pami} and high-speed obstacle avoidance~\cite{Falanga20Science}.
Recently, event cameras have been increasingly applied in automotive settings~\cite{Perot20nips,Maqueda18cvpr,muglikar213dv}, where they promise to solve computer vision tasks in challenging edge-case scenarios, such as when exiting a tunnel into bright sunlight~\todo{\cite{Gehrig21ral,Rebecq19pami}} or when children unexpectedly jump in front of a car.

For the latter, extracting detailed and dense semantic information is essential for any automotive safety system.
In particular, event-based semantic segmentation promises to significantly improve the reliability and safety of these systems by leveraging the robustness to lighting conditions and the low latency of event cameras.
However, due to the novelty of the sensor, event-based semantic segmentation is still in its infancy, resulting in a lack of high-quality event-based semantic segmentation datasets.
While some datasets exist~\cite{Alonso19cvprw,RAL21Gehrig}, these are either synthetic or feature pseudo labels, which are produced by an image-based network running on low-quality grayscale images. 
As a result, methods trained on these datasets typically exhibit suboptimal performance~\cite{Wang21cvpr,Alonso19cvprw}.

\input{floats/fig_eye_catcher.tex}

In this work, we make significant strides toward high-quality event-based semantic segmentation by addressing the above limitation on two fronts: First, we generate a new event-based semantic segmentation dataset, named DSEC-Semantic, based on the stereo event camera dataset for driving scenarios (DSEC)~\cite{Gehrig21ral}. 
The labels are generated via pseudo-labeling on high-quality RGB images and filtered by manual inspection. 
Second, we introduce ESS, a novel unsupervised domain adaptation (UDA) method specifically tailored to event data, which transfers a task from a labeled image dataset to an unlabeled event domain, see Fig.~\ref{fig:eye_catcher}. 
Compared to other methods, it does not require video data~\cite{Gehrig20cvpr} or per-pixel paired events and images~\cite{Yuhunang20eccv,Wang21cvpr} and does not need to hallucinate motion-imbued events from still images~\cite{Messikommer22ral}. 
In fact, generating events from single images remains an ill-posed problem that so far has only been studied via adversarial learning, which is prone to mode collapse.
Instead, our method produces a recurrent, motion-invariant event embedding, which is aligned with image embeddings during the training process, facilitating the transfer between domains.

We perform extensive evaluation both on the existing DDD17~\cite{Binas17icml,Alonso19cvprw} benchmark and our new DSEC-Semantic benchmark.
On DDD17, we report a \todo{6.98\%} higher mean intersection over union (mIoU) compared to other UDA methods, and when using additional event labels, ESS outperforms supervised methods by 2.57\%.
On DSEC-Semantic, we show a \todo{4.17\%} higher mIoU when compared to other UDA approaches. 
Additionally, when combined with supervised learning, our method achieves 1.53\% higher mIoU than other state-of-the-art supervised methods. 
Our contributions can be summarized as follows:
\begin{enumerate}
    \item We present a UDA method that leverages image datasets to train neural networks for event data. It does this by directly aligning recurrent, motion-invariant event embeddings with image embedding without requiring paired data, video, or explicit event generation.
    \item We show that our method outperforms existing state-of-the-art UDA and supervised methods both on an existing and our newly introduced benchmark.
    \item We contribute a new high-quality dataset for event-based semantic segmentation, based on high-quality RGB frames from the large-scale DSEC dataset.
\end{enumerate}
Finally, since ESS is general-purpose, it unlocks the virtually unlimited supply of existing image datasets, thereby democratizing them for event camera research. 
These datasets will pave the way for new and exciting research directions in new fields which were previously inaccessible for event cameras.

%% file: floats/fig_eye_catcher.tex
\begin{figure}[t!]
    \centering
    \begin{tabular}{c}
         \includegraphics[width=0.99\textwidth]{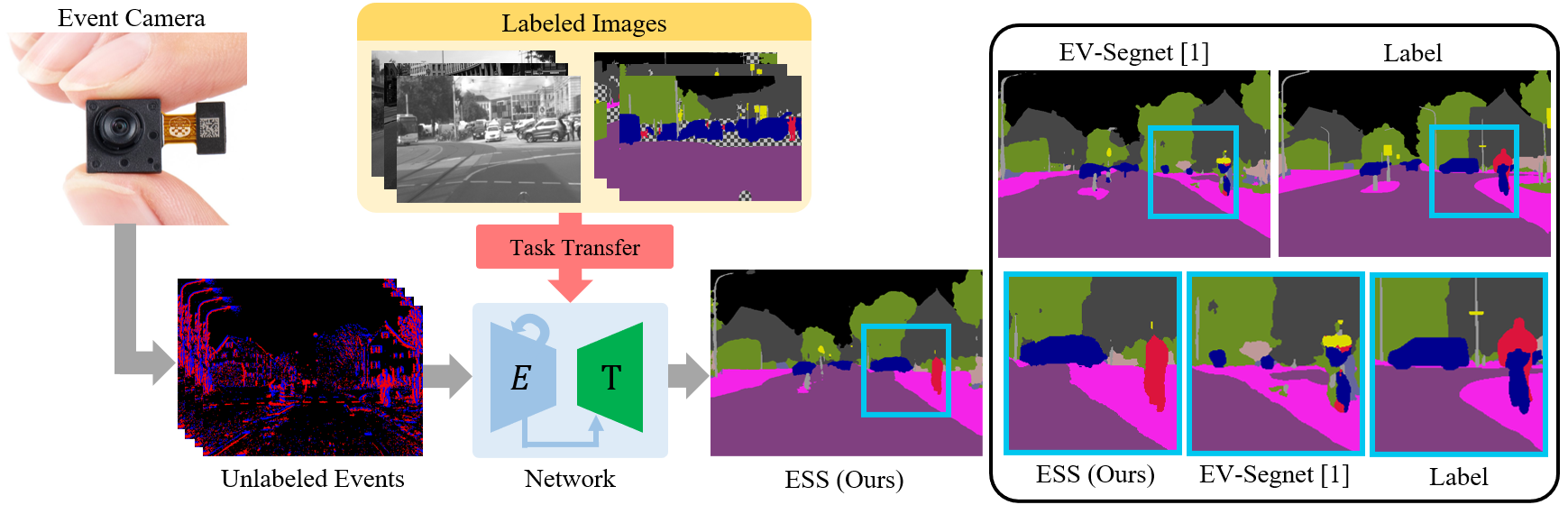}
    \end{tabular}
    \vspace{-2ex}
    \caption{
    In this work, we present ESS, a method for fine-grained event-based semantic segmentation. 
    Due to the novelty of the sensor, only few datasets for event cameras are available. 
    For this reason, we leverage existing image-based datasets for training neural networks for events using a novel UDA approach. 
    Compared to existing methods, our approach does not require video or paired data and does not need to hallucinate motion to construct events. 
    Our method can detect fine-grained objects, such as cars and pedestrians, outperforming state-of-the-art methods in UDA and supervised settings, on a common benchmark and our newly created benchmark with high-quality labels (right).
    }
    \label{fig:eye_catcher}
\end{figure}

%% file: sections/related_work.tex
\section{Related Work}
\subsection{Event-based Semantic Segmentation}
The first work to use events for the task of semantic segmentation was~\cite{Alonso19cvprw}, which also introduced the first event-based semantic segmentation dataset based on the driving dataset DDD17~\cite{Binas17icml}. 
It used an Xception-type network~\cite{Chollet17cvpr} to show robust performance in edge-case scenarios, where standard images are overexposed.
The semantic labels (also known as pseudo-labels) on DDD17 were generated by a pre-trained network running on the grayscale frames of the DAVIS346B~\cite{Brandli14ssc}. 
This sensor features per-pixel aligned events and frames and has been useful for a variety of domain adaptation works. 
However, it has a low resolution and poor image quality, which results in significant artifacts in the resulting pseudo labels. 
Additionally, due to the low resolution of the DAVIS346B, multiple classes need to be merged, reducing the granularity of the labels.
In parallel, the simulated EventScape dataset~\cite{Gehrig21ral,Hidalgo20threedv} includes high-quality semantic labels but was recorded in the CARLA simulator~\cite{Dosovitskiy17corl}, and thus exhibits a sim-to-real gap.

Follow-up work by~\cite{Gehrig20cvpr} improved on the results of \cite{Alonso19cvprw} by leveraging additional labeled video datasets for events by augmenting training data with synthetic events converted from video. 
While this method allows networks to be trained on synthetic and real events resulting in a significant performance boost, it requires the availability of video datasets, which are not as common as datasets containing still images and are especially rare for semantic segmentation. 
For this reason,~\cite{Wang21cvpr} combines labeled image datasets such as Cityscape~\cite{Cordts16cvpr} with unlabeled events and frames from a DAVIS to decrease the dependence on video data. 
In fact, they report an increase in the semantic segmentation performance but still rely on per-pixel paired data from a DAVIS for successful transfer.
Another supervised semantic segmentation method~\cite{Wang21iccv} leverages the event-to-image transfer to help with the task of semantic segmentation. 
However, they rely on a labeled event dataset and do not consider recurrent event embeddings.
Our method also leverages UDA for event-based semantic segmentation but differs from existing work in a few key points: \emph{(i)} it only leverages datasets of still images,  \emph{(ii)} does not require per-pixel paired events and frames, and \emph{(iii)} uses a recurrent network to generate motion-invariant event embeddings.
Since UDA methods have become instrumental for event-based semantic segmentation, we review them next.
\subsection{Unsupervised Domain Adaptation}
A common challenge for novel sensors such as event cameras is the lack of labeled datasets.
To tackle this challenge, multiple works try to leverage labeled images to train networks for event cameras.
This transfer from a labeled source domain (images) to an unlabeled target domain (events) is generally defined as Unsupervised Domain Adaption (UDA).

Event-to-image reconstruction methods~\cite{Rebecq19pami,Reinbacher16bmvc,Bardow16cvpr} were the first to address this setting. 
Most recently, E2VID~\cite{Rebecq19pami} uses a recurrent network to convert events to video, which can then be processed with standard image-based networks.
It, however, requires the overhead of converting events first to images and does not leverage unlabeled target events to help the task transfer from images to events.
Instead of going from events to frames, VID2E converts labeled video sequences to event sequences.
The synthetic events can then be used to train a network on the corresponding image labels, which transfers to real events.
While this reduces the overhead of needing to convert events to video, it still cannot adapt to an unlabeled event domain. Moreover, it requires labeled video datasets, which excludes the majority of existing image datasets.

One of the first approaches to do explicit domain adaptation was network grafting~\cite{Yuhunang20eccv}, which replaces the encoder of a pre-trained image network with an event encoder, and finetuning it with paired events and images. 
However, it needs to be trained with a consistency loss, which requires paired event and image data, and is thus not be applicable in the UDA setting. 
Moreover, this constraint limits the kind of datasets that can be leveraged to those recorded with per-pixel aligned events and frames, which excludes most existing image datasets.
EvDistill~\cite{Wang21cvpr} lifted this limitation by instead leveraging unpaired events and images, with unlabeled events and labeled images. While this approach could transfer from unpaired Cityscapes labels to events from DDD17, they only report the segmentation performance based on paired images and events. Strictly speaking, this can thus not be considered as a UDA method.
Instead, a pure UDA method for image-to-event transfer is \cite{Messikommer22ral}, which splits the embedding space into motion-specific features and features shared by both image and events.
They use adversarial learning to align image and event embedding spaces for the task of classification and object detection.
However, this approach relies on generating fake events, which requires the hallucination of motion from still images.
This hallucination is ill-posed and thus hinders the feature alignment, which is crucial for a successful task transfer from images to events.
Our method, ESS, addresses these limitations by transferring from single images to events without the need for hallucinating motion.
This task transfer is achieved by generating motion-invariant event embeddings, leveraging the pre-trained E2VID~\cite{Rebecq19pami} encoder which are then aligned with the embedding space of single images via a dedicated image encoder.
Since the resulting event embeddings do not contain motion information, they can be easily aligned in the embedding space, facilitating task transfer.

%% file: sections/method.tex
\section{Approach}
\label{sec:approach}

\input{floats/fig_method_overview}
 
Our method transfers a task from a labeled source domain $\mathbb{I}=\{(I_i,\mathcal{L}_i)\}_{i=1}^M$ to an unlabeled target domain $\mathbb{E}=\{\mathcal{E}_i\}_{i=1}^N$. 
More specifically, the source domain $\mathbb{I}$ consists of images $I_i\in\mathbb{R}^{H\times W}$ and labels in form of semantic maps $\mathcal{L}\in \mathbb{Z}^{H\times W}_c$, where $c$ is the number of classes. 
The event domain  $\mathbb{E}$ consists of data recorded by an event camera. 
Event cameras have independent pixels which trigger each time the log brightness changes by a fixed threshold. 
The resulting data is an asynchronous stream,
$\mathcal{E}_i=\{e_{i,j}\}_{j=1}^{n_i}$ made up of temporally ordered events $e_{i,j}$, each encoding the pixel coordinate $\mathbf{x}_{i,j}$, timestamp with microsecond-level resolution $t_{i,j}$ and polarity $p_{i,j}\in \{-1,1\}$ of the brightness change. For more information about the working principles of event cameras, see~\cite{Gallego20pami}.

The goal of our approach is to train a neural network $F$ which takes event sequences\footnote{For clarity, we omit the subscript i in the future.} $\mathcal{E}$ as input and outputs the task variable in form of pixel-wise semantic predictions $\mathcal{L}$. 
At training time, it only has access to image labels from the source domain $\mathbb{I}$, but can leverage unlabeled events from the target domain $\mathbb{E}$. 

An overview of our method is shown in Fig. \ref{fig:method_overview}. 
Our method works by first encoding events into a motion-invariant embedding $\textbf{z}_\text{event}$ using the E2VID~\cite{Rebecq19pami} encoder $E_{\text{E2VID}}$ and decoding these to an image reconstruction using the decoder $D_\text{E2VID}$. 
\todo{This event embedding preserves sufficient semantic information for the segmentation task but excludes motion information since it is used to reconstruct motion-invariant still images.
The image reconstruction and events then formulate a pseudo pair in the source and target domain, which can be leveraged to align the embedding space.
}
Consequently, we use an image encoder $E_\text{img}$ to approximate \todo{the motion-invariant} embedding.
Finally, a \todo{shared} task network $T$ generates task predictions from image and event embeddings.

\subsection{Network Overview}
\label{sec:approach_network_overview}

In a first step, we convert an event stream $\mathcal{E}$ to a sequence of grid-like representations~\cite{Gehrig19iccv}, such as \emph{voxel grids}~\cite{Zhu19cvpr} $\mathbf{V}_k$. 
Each voxel grid is constructed from non-overlapping windows with a fixed number of events, see supplementary for more details.
Next, we produce a recurrent, multi-scale embedding $\mathbf{z}_\text{event}$, with 
\begin{equation}
\mathbf{z}^k_\text{event}=E_\text{E2VID}(\mathbf{V}_k, \mathbf{z}^{k-1}_{\text{event}}), \quad k=1,...,N,
\end{equation}
and $\textbf{z}_\text{event}=\textbf{z}^N_\text{event}$. 
Simultaneously, we train an image encoder $E_\text{img}$ which produces image embeddings $\mathbf{z}_{\text{img}}~=~E_\text{img}(I)$. 
These embeddings are used in three branches of the training framework, see Fig.~\ref{fig:method_overview}. 
First, we use the image and event embeddings to produce a task prediction via a task network $T$
\begin{align}
\mathcal{L}_\text{img}=T(\mathbf{z}_\text{img}) \quad \text{ and } \quad \mathcal{L}_\text{event}=T(\mathbf{z}_\text{event}),
\end{align}
with $\mathcal{L}_{\text{img/event}}\in\mathbb{R}^{H\times W\times c}$. 
Second, we also use the event embedding to generate an image reconstruction via the decoder $D_\text{E2VID}$, as $\hat I = D_\text{E2VID}(\mathbf{z}_\text{event})$\todo{, which results in a pseudo pair $(\hat I, \mathcal{E})$ in the source and target domain}. 
Finally, $E_\text{img}$ reencodes the resulting image and produces a task prediction 
 \begin{align}
     \hat{\mathbf{z}}_\text{img}=G\left(D_\text{E2VID}(\textbf{z}_\text{event})\right) \quad \text{ and }\quad \hat{\mathcal{L}}_{\text{img}}=T(\hat{\mathbf{z}}_\text{img}).
 \end{align}
\todo{Details} of $D_\text{E2VID}, E_\text{E2VID}, E_\text{img}$ and $T$ are given in the supplementary.
\todo{In the following, we explain how the alignment of the motion-invariant embeddings is enforced by multiple consistency losses.
This alignment is crucial since it ensures that the task decoder $T$ can be applied in the event and image domain.
}

\todo{\subsection{Aligning Motion-Invariant Embedding}
\label{sec:approach_aligning_motion_invariant_embedding}
With pseudo pairs $(\hat I, \mathcal{E})$ in the source and target domain, our method leverages several \textbf{consistency losses} to align the motion-invariant embeddings.
Inspired by prior works~\cite{Messikommer22ral,Wang21cvpr}, we enforce an alignment between event embeddings and reencoded event embeddings via an $L_1$ distance and between task predictions via the symmetric Jensen-Shannon divergence
 \begin{align}
 L_\text{cons. emb.} &= \Vert \textbf{z}_{\text{event}}-\hat{\textbf{z}}_{\text{img}} \Vert_{1}\\
 L_\text{cons. pred.} &= \frac{1}{2} D_{\text{KL}}(T(\mathbf{z}_\text{event}) \Vert T\left(\hat{\mathbf{z}}_\text{img})\right) + 
                        \frac{1}{2} D_{\text{KL}}(T\left(\hat{\mathbf{z}}_\text{img})\right \Vert T(\mathbf{z}_\text{event})).
 \end{align}
 Furthermore, we tighten the alignment by minimizing the $L_1$ distance between intermediate features $T^{(i)}(\mathbf{z}_\text{event})$ and $T^{(i)}(\hat{\mathbf{z}}_\text{img})$ produced while decoding the embeddings $\mathbf{z}_\text{event}$ and $\hat{\mathbf{z}}_\text{img}$ resulting in the following loss 
 \begin{equation}
     L_{\text{cons. task}} = \sum_i\left\Vert T^{(i)}(\hat{\mathbf{z}}_\text{img}) - T^{(i)}(\mathbf{z}_\text{event}) \right\Vert_1.
 \end{equation}
 }
 
 \subsection{Losses and Optimization}
 \label{sec:approach_losses_and_optimization}
\todo{At each training step, we additionally compute the task loss in form of the cross-entropy and Dice loss in the image domain by leveraging the available labels} $\mathcal{L}$, 
 \begin{equation}
     L_{\text{task}} = \text{CrossEntropy}(T(\mathbf{z}_\text{img}), \mathcal{L}) + \text{Dice}(T(\mathbf{z}_\text{img}), \mathcal{L}).
 \end{equation}
\todo{Finally, we sum up the task loss and the consistency losses}
  \begin{equation}
     L_\text{total} = \lambda _{1}L_{\text{task}} + \lambda _{2}L_\text{cons. emb.} + \lambda _{3}L_\text{cons. pred.} + \lambda _{4}L_\text{cons. task},
 \end{equation}
\todo{where $\lambda _{1}$, $\lambda _{2}$, $\lambda _{3}$, and $\lambda _{4}$ are the hyper-parameters.}
 
\noindent\textbf{Optimization}
\todo{We perform a two-stage network gradient accumulation for each optimization step, shown in Fig. \ref{fig:method_overview}. 
During the first stage (left), we use an image and label pair to compute the task loss, which we only use to update the network gradients of the task decoder $T$. 
During the second stage (right), we train on unlabeled events. Here, we freeze the E2VID encoder/decoder pair and the task network in the second branch (Fig. \ref{fig:method_overview}, top right). 
After computing the consistency losses, we accumulate the gradients for the image encoder $E_\text{img}$ and the task decoder $T$ in the first branch. 
We perform one parameter update step on the network after accumulating these gradients.}

%% file: floats/fig_method_overview.tex
\begin{figure}[t!]
\centering
\includegraphics[width=0.99\textwidth]{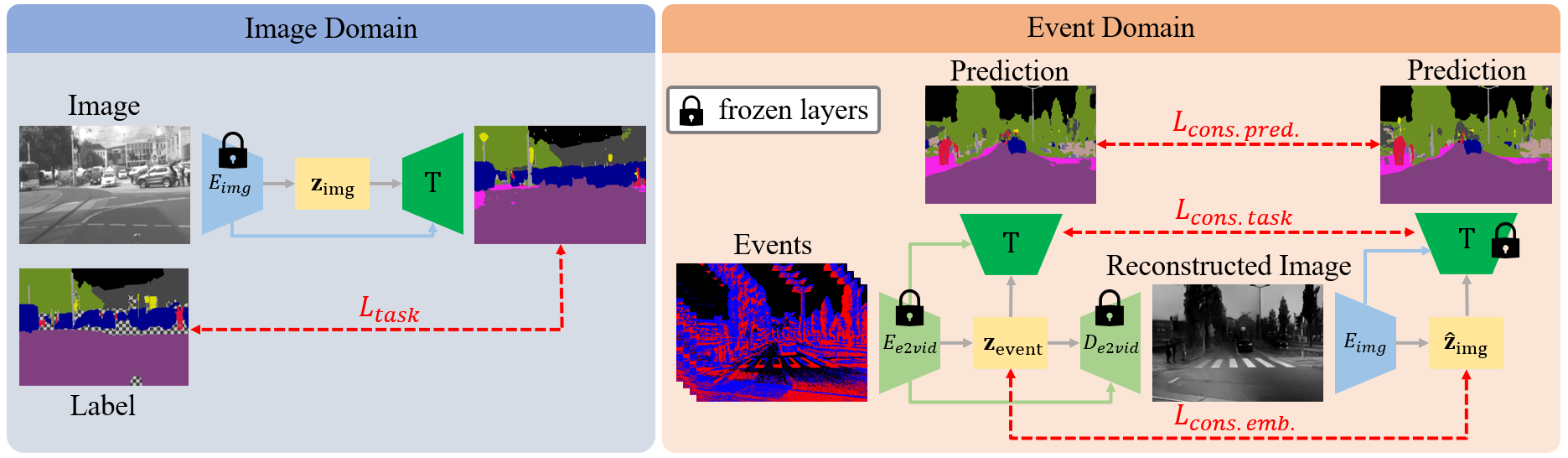}
\vspace{-1ex}
\caption{Our method ESS, performs unsupervised domain adaptation by leveraging labeled image datasets (source domain, left) to train networks for event cameras in an unlabeled target domain (right). 
In the source domain, it performs supervised learning \todo{on the task network while not training the image encoder}.
In the event-domain, it uses the recurrent E2VID encoder to produce motion-invariant event embeddings, which are decoded and reencoded using the image encoder. Various consistency losses align these embeddings, forcing the image encoder to behave similarly to the event encoder. In this stage, both task and image encoder is trained. At test-time ESS simply uses the E2VID encoder and task decoder for prediction, and thus remains lightweight.}
\label{fig:method_overview}
\end{figure} 

%% file: sections/results.tex
\newcolumntype{C}[1]{>{\centering}m{#1}}

\section{Experiments}
We start off in Sec.~\ref{sec:ddd17}, by validating our method on the commonly used DDD17 benchmark~\cite{Alonso19cvprw}, where we compare against supervised~\cite{Alonso19cvprw,Wang21iccv}, pixel-wise paired~\cite{Wang21cvpr}, and UDA methods~\cite{Rebecq19pami,Messikommer22ral,Gehrig20cvpr}. We then introduce our newly generated DSEC-Semantic dataset in Sec.~\ref{sec:dsec}, which contains higher quality semantic labels, and report comparisons on this dataset.
Finally, in Sec.~\ref{sec:ablation} we perform ablation studies to verify the effectiveness of the proposed design choices.
For more results, we refer to the supplementary.

\noindent\textbf{Baseline Methods}
We compare our task transfer method with the two state-of-the-art UDA approaches E2VID~\cite{Rebecq19pami} and EV-Transfer~\cite{Messikommer22ral}.
For E2VID, we take the pre-trained network weights provided by the authors to convert events to grayscale images. 
We retrain a semantic segmentation network on labeled grayscale images from Cityscapes~\cite{Cordts16cvpr}, which we then apply to the reconstructed images. 
E2VID is indeed a UDA method since it does not require a labeled event dataset nor paired image and event data. 
However, different from our method, it cannot be retrained for a specific target domain, performing zero-shot UDA.

In contrast, EV-Transfer~\cite{Messikommer22ral} leverages unlabeled targets events for classification and object detection. 
To adapt the open-source implementation to semantic segmentation, we add the same task decoder as our method without skip connections.
We report DDD17 results for EV-Distill~\cite{Wang21cvpr} and DTL~\cite{Wang21iccv}, but do not include them on DSEC-Semantic since open-source training code is not available, and they require paired images and events, which are not available on that dataset.
Finally, we compare against the supervised methods VID2E~\cite{Gehrig20cvpr} and EV-Segnet~\cite{Alonso19cvprw}, which we retrain on DSEC-Semantic based on open-source code.

\subsection{DDD17 for Semantic Segmentation}
\label{sec:ddd17}
The DAVIS Driving Dataset (DDD17) for semantic segmentation targets automotive scenarios and contains 12 hours of driving data recorded with a DAVIS~\cite{Brandli14ssc}, which provides per-pixel aligned and temporally synchronized events and gray-scale frames. 
In \cite{Alonso19cvprw}, they used a pre-trained Xception network~\cite{Chollet17cvpr} to generate semantic pseudo-labels on the DAVIS frames. Since the DAVIS only features a low resolution, they fuse several classes and only provide labels for 6 merged classes: flat (road and pavement), background (construction and sky), object, vegetation, human, and vehicle. 
In this section, we will compare our method against related work in two settings: \emph{(i)} in the UDA setting, where we only use unlabeled events, labeled frames and present them to the network in an unpaired fashion, and \emph{(ii)} in a paired event and frame setting as well as in the supervised setting, where we introduce additional labels in the event-domain.

\noindent\textbf{Implementation Details}
We use Cityscapes~\cite{Cordts16cvpr} as the labeled source domain and DDD17 as the unlabeled target domain. 
\todo{The hyper-parameters $\lambda _{1}$, $\lambda _{2}$, $\lambda _{3}$, and $\lambda _{4}$ are set as 1, 0.01, 1, and 0.01, respectively.
We set the learning rates as $1\times 10^{-5}$ for $E_\text{img}$ and $1\times 10^{-4}$ for $T$.
We empirically found that having a smaller learning rate on $E_\text{img}$ and activating the accumulation of gradients for $E_\text{img}$ in the first stage help improve the results.}
We train our model using the RAdam optimizer~\cite{Liu18iclr} with a batch-size of 16 for 50'000 iterations.    
Additionally, for the comparison with E2VID~\cite{Rebecq19pami} in the UDA setting, we retrain the image encoder and task network (forming a U-Net) on grayscale images and labels from the Cityscapes dataset~\cite{Cordts16cvpr}. 
Similar to our method, we train \cite{Messikommer22ral} in our UDA setting with the same source and target domains.
As commonly done, we report the accuracy as well as the mean intersection over union (mIoU) on the resulting segmentation maps, which better highlights the accuracy on smaller objects.

\input{floats/table_ddd17_uda}
\input{floats/table_ddd17_supervised}

\noindent\textbf{UDA Comparison}
Tab.~\ref{tab:exp_ddd17_uda} shows that our method outperforms the runner-up VID2E by a large margin of \todo{6.98\%} mIoU. 
VID2E converts DDD17 grayscale images to events and trains on the DDD17 labels. 
However, it suffers from a domain gap between synthetic and real events, which it cannot bridge using domain adaptation. 
Similarly, E2VID~\cite{Rebecq19pami} cannot perform domain adaptation, which is why it achieves a lower performance.
EV-Transfer~\cite{Messikommer22ral} does domain adaptation but is still outperformed by our method. This is because we use a recurrent event encoder, which retains memory and can thus handle static scenes, which do not trigger events, leading to better predictions. Moreover, since our method aligns motion-invariant event embeddings, it does not rely on adversarial training and is therefore much simpler to train.
Fig.~\ref{fig:ddd17_uda_samples} shows qualitative results of the tested methods.

\input{floats/fig_ddd17_uda_samples}
\input{floats/fig_ddd17_ours_vs_labels}

\noindent\textbf{State-of-the-art Comparison}
Here, we show that, when combined with supervised learning, our method outperforms state-of-the-art methods. To do this, we add an additional task loss during training at the first task branch (Fig.~\ref{fig:method_overview}, top right), which allows our method to simultaneously leverage image and event labels.
We also compare against supervised methods DTL~\cite{Wang21iccv} and EV-Distill~\cite{Wang20cvpr}, which rely on the paired images and events provided by the DAVIS. 
We report results for two variations of our approach: The first is trained using only the recurrent encoder and task decoder, in a supervised setting using labeled events (Fig.~\ref{fig:method_overview}, event domain, top left). The second combines supervised training on events with our full domain adaptation framework, including labeled images for improved performance. These methods are labeled with ``events" and ``events+frames" respectively in Tab. \ref{tab:exp_ddd17_supervised}.
As reported in Tab.~\ref{tab:exp_ddd17_supervised}, our method outperforms the runner-up DTL by 2.57\% mIoU if trained in a supervised setting with events.
DTL is a feed-forward network, which shows that our recurrent encoder boosts performance, especially in the near static scenes of DDD17.
An additional advantage of our method compared to standard supervised methods is that it can leverage image labels in combination with event labels.
From the Tab. \ref{tab:exp_ddd17_supervised}, it can be observed that the additional image labels do not lead to a performance improvement.
In fact, this can be explained by the fact that DDD17 semantic labels are not always accurate. In several examples (see Fig.~\ref{fig:ours_vs_labels}), we found that our method predicted objects which were not present in the labels, but were clearly visible in the images and thus reduced the segmentation performance. 
Fig.~\ref{fig:ours_vs_labels} shows that our method trained supervised on events and images sometimes provides more accurate predictions than the pseudo-labels from DDD17.

\input{floats/fig_dataset_overview}

\subsection{DSEC-Semantic}
\label{sec:dsec}
The semantic segmentation labels for DDD17 suffer from artifacts caused by the low-quality and low-resolution grayscale images, shown in Fig.~\ref{fig:dataset_overview}.
For this reason, we generate a new semantic segmentation dataset based on DSEC~\cite{Gehrig21ral}.
DSEC contains 53 driving sequences collected in a variety of urban and rural environments in Switzerland and was recorded with automotive-grade standard cameras and high-resolution event cameras.
We use the pseudo labeling scheme adopted in \cite{Alonso19cvprw} with the high-quality images provided by the left color FLIR Blackfly S USB3 with a resolution of 1440 $\times$ 1080.
\todo{The semantic labels are generated by first warping the images from the left frame-based camera to the view of the left monochrome Prophesse Gen3.1 event camera with a resolution of 640 $\times$ 480.
We then apply a state-of-the-art semantic segmentation method~\cite{Tao20arxiv} to the warped images to generate the labels.}
By doing so, we obtain fine-grained labels for 19 classes, which we convert to 11 classes: background, building, fence, person, pole, road, sidewalk, vegetation, car, wall, and traffic sign.
Since frame cameras suffer from image degradation in challenging illumination scenes, we only label the sequences recorded during the day, \todo{which results in 8082 labeled frames for the training and 2809 labeled frames for the test split.}
For more details, we refer to the supplementary.
Compared to labels from DDD17, our labels feature much higher quality and more details, as can be observed in Fig.~\ref{fig:dataset_overview}.
We believe that our generated semantic labels can also spur future work in multi-modal semantic segmentation as the DSEC dataset includes measurements of a LiDAR, one frame-based, and one event-based stereo-camera pair.

\noindent\textbf{Implementation Details}
Similar to the experiments on DDD17, we leverage the Cityscapes datasets as the labeled source dataset.
\todo{
The difference is that we use the DSEC-Semantic dataset as the target domain. 
The hyper-parameters $\lambda _{1}$, $\lambda _{2}$, $\lambda _{3}$, and $\lambda _{4}$ are now set as 1, 1, 1, and 1, respectively.
We use the same RAdam optimizer to train our model with a larger learning rate of $5\times 10^{-4}$ (for both $E_\text{img}$ and $T$), and a smaller batch-size of 8, for 25'000 iterations.}

\noindent\textbf{UDA Comparison}
In this setting, we compare against the UDA methods \cite{Rebecq19pami,Messikommer22ral}, which can deal with unpaired, labeled image and unlabeled event data.
As can be observed in Tab.~\ref{tab:exp_dsec_uda}, our method outperforms state-of-the-art UDA methods by a margin of \todo{4.17\%} mIoU.
Again, our method benefits from a recurrent architecture, and a simpler training regime that does not rely on adversarial training. 
Moreover, it can be adapted to the target domain, showing a large gap to methods that cannot do so, such as~\cite{Rebecq19pami}.
Fig.~\ref{fig:dsec_uda_samples} shows qualitative examples verifying the benefits of our method.
\input{floats/fig_dsec_uda_samples}
\input{floats/table_dsec_uda}

\textbf{State-of-the-art Comparison}
Here, we adopt the same setting as for the DDD17, where we train our method with a supervised task loss on training labels in the event domain. 
See Fig.~\ref{fig:dsec_supervised_samples} for qualitative samples.
In this supervised setting, we compare against EV-Segnet~\cite{Alonso19cvprw}.
Additionally, we also provide results for our method using both image and event labels during training.
Without considering the image labels in the training, our method achieves a performance comparable with EV-Segnet with a higher accuracy score but a slightly lower mIoU, see Tab.~\ref{tab:exp_dsec_supervised}.
However, if we use the full potential of our method by using the image labels as well, we achieve state-of-the-art performance on DSEC-Semantic, outperforming EV-SegNet by 1.53\% mIoU.

\input{floats/fig_dsec_supervised_samples}
\input{floats/table_dsec_supervised}

\subsection{Ablation Studies}
\label{sec:ablation}
\textbf{Loss importance} To verify the effectiveness of the proposed framework, we ablate the introduced loss functions by removing them during training.
Tab.~\ref{tab:exp_dsec_ablation} reports the results of those experiments on DSEC-Semantic for the UDA setting.
It can be observed that omitting the consistency loss in the embedding space, $L_\text{cons. emb.}$, leads to a \todo{5.56\%} drop in mIoU, showing its importance to align the embedding spaces.
Similarly, omitting $L_\text{cons. pred.}$ leads to a \todo{1.28\%}, and omitting $L_\text{cons. task.}$ leads to a \todo{2.09\%} drop, highlighting the importance of both.

\input{floats/table_combined_ablation}

\noindent\textbf{Embedding Alignment}
The studied UDA methods operate by aligning events and frame embeddings. 
For E2VID, these embeddings are image reconstructions and images, for EV-Transfer and our work, these are image and event embeddings.
To study this alignment, we perform the following comparison: On DSEC-Semantic, we construct pairs of event and image embeddings, which we each decode to the logits of the semantic map, $T(\mathbf{z}_\text{event})$ and $T(\mathbf{z}_\text{img})$. 
While for EV-Transfer and our method, we use the dedicated task network to decode these embeddings, for the E2VID-baseline, we use the network trained on Cityscapes on both images and image reconstructions, to construct paired predictions. 
We then measure the consistency of these maps across pairs, via the symmetric KL divergence, which we report in Tab.~\ref{tab:exp_ddd17_alignment}. As can be seen, our approach has a three times lower KL divergence with 0.025, than the runner-up E2VID with 0.073. This indicates that our method aligns image and event embeddings better than other methods, facilitating domain transfer.

%% file: floats/table_ddd17_uda.tex
\begin{table}[t!]
\caption{
Performance of EV-Transfer, E2VID, VID2E, and our method on DDD17 in the UDA setting, in which the labels of Cityscapes and unlabeled events of DDD17 are available.
\todo{Results report the mean and standard deviation of 3 runs with different random seeds except for the VID2E method which is taken from ~\cite{Gehrig20cvpr}.}}
\centering
\scalebox{1}{
\begin{tabular}{m{2.7cm}C{2.7cm}>{\centering\arraybackslash}m{2.7cm}}
\hline
Method  & Accuracy [\%] $\uparrow$ & mIoU [\%] $\uparrow$ \\
 \hline
EV-Transfer~\cite{Messikommer22ral} & \todo{47.37$\pm$4.53} & \todo{14.91$\pm$0.61} \\
E2VID~\cite{Rebecq19pami}           & \todo{83.24$\pm$2.60} & \todo{44.77$\pm$3.70} \\
VID2E~\cite{Gehrig20cvpr}           & 85.93 & 45.48 \\
\textbf{ESS (ours)}                                & \todo{\textbf{87.86$\pm$0.57}} & \todo{\textbf{52.46$\pm$0.63}}  \\ 
\hline
\end{tabular}
}
\label{tab:exp_ddd17_uda}
\end{table}


%% file: floats/table_ddd17_supervised.tex
\begin{table}[t!]
\caption{
Results on DDD17 in the setting in which all of the available training data can be used.
That includes real events with corresponding labels (events) and the possible combination with either synthetic events based on grayscale images of DDD17 (synthetic+events) or image labels (events+frames).
}
\centering
\scalebox{1}{
\begin{tabular}{m{2.7cm}C{2.7cm}C{2.7cm}>{\centering\arraybackslash}m{2.7cm}}
\hline
Method  & Training Data & Accuracy [\%] $\uparrow$ & mIoU [\%] $\uparrow$ \\
 \hline
EVDistill~\cite{Wang21cvpr}    & events             &   -   & 58.02 \\
EV-SegNet~\cite{Alonso19cvprw} & events             & 89.76 & 54.81 \\
VID2E~\cite{Gehrig20cvpr}      & synthetic+events   & 90.19 & 56.01 \\
DTL ~\cite{Wang21iccv}         & events             &   -   & 58.80 \\
\textbf{ESS (ours)}            & events             & \textbf{91.08} & \textbf{61.37}  \\
\textbf{ESS (ours)}            & events+frames      & 90.37 & 60.43  \\
\hline
\end{tabular}
}
\label{tab:exp_ddd17_supervised}
\end{table}

%% file: floats/fig_ddd17_uda_samples.tex
\begin{figure}[t!]
\centering
\includegraphics[width=0.99\textwidth]{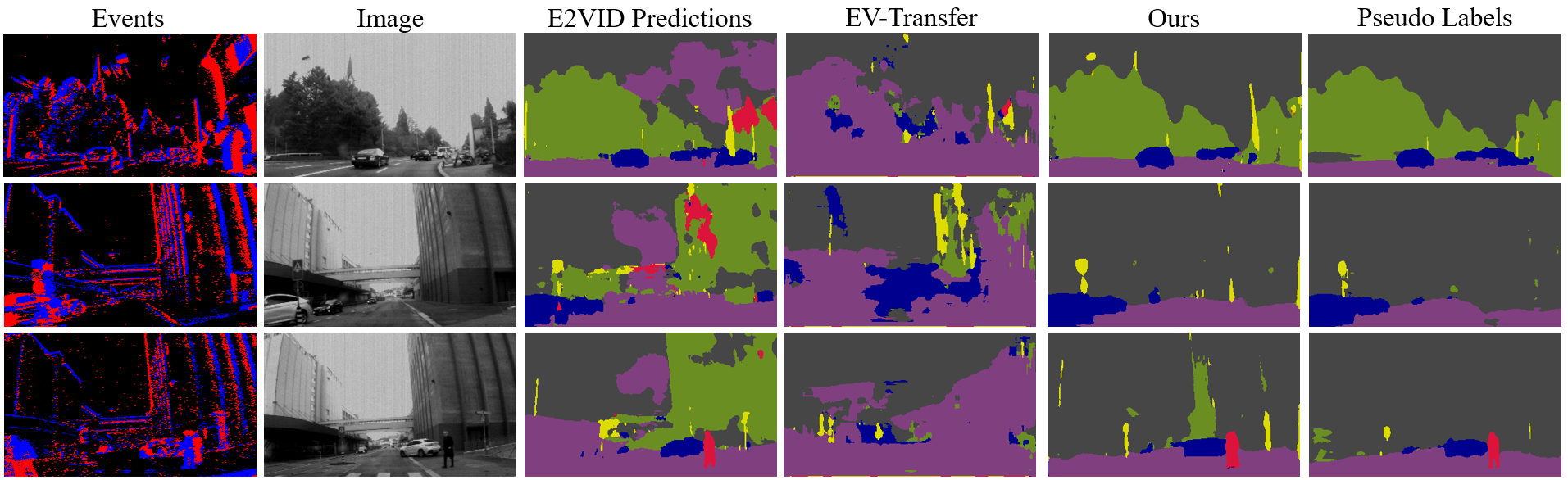}
\caption{
Qualitative samples on DDD17 for the UDA setting, i.e., no event labels are available during training.
Compared to EV-Transfer and E2VID, our method can more reliably predict smaller details such as people.
}
\label{fig:ddd17_uda_samples}
\end{figure} 

%% file: floats/fig_ddd17_ours_vs_labels.tex
\begin{figure}
    \includegraphics[width=0.99\textwidth]{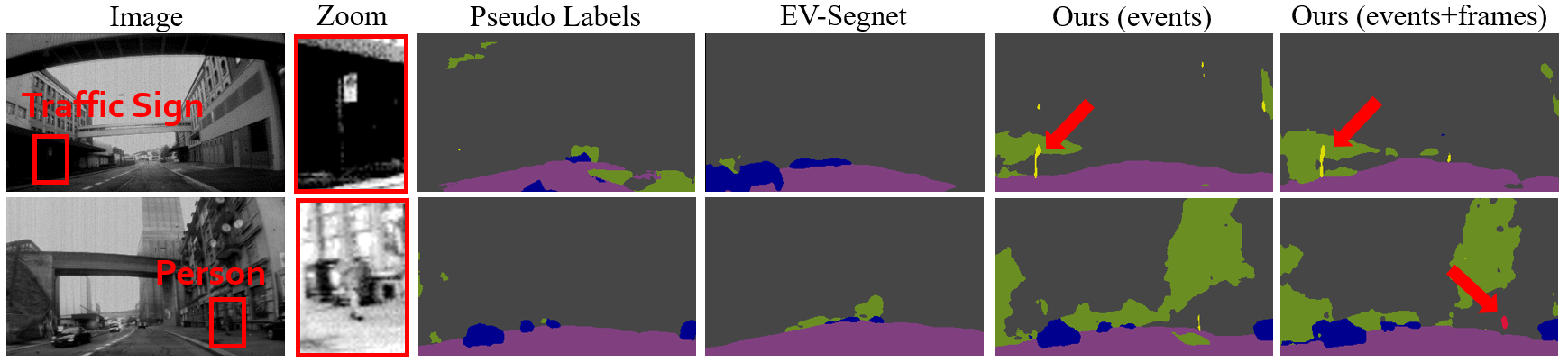}
    \caption{
    Predictions of EV-Segnet and our method trained once purely with event labels (events) and once also with image labels (events+frames).
    Due to the low-quality of the DDD17 semantic labels, small objects are sometimes missed in the pseudo labels, (zoomed-in and brightened image patch in the red box).
    These objects are more reliable detected if our method is trained on the high-quality labels of Cityscapes.
    This can lead to a lower detection score on DDD17 even though the predictions of our method trained on events and frames provide more finegrained detections.
    } 
    \label{fig:ours_vs_labels}
\end{figure}

%% file: floats/fig_dataset_overview.tex
\begin{figure}[t!]
\centering
\includegraphics[width=0.99\textwidth]{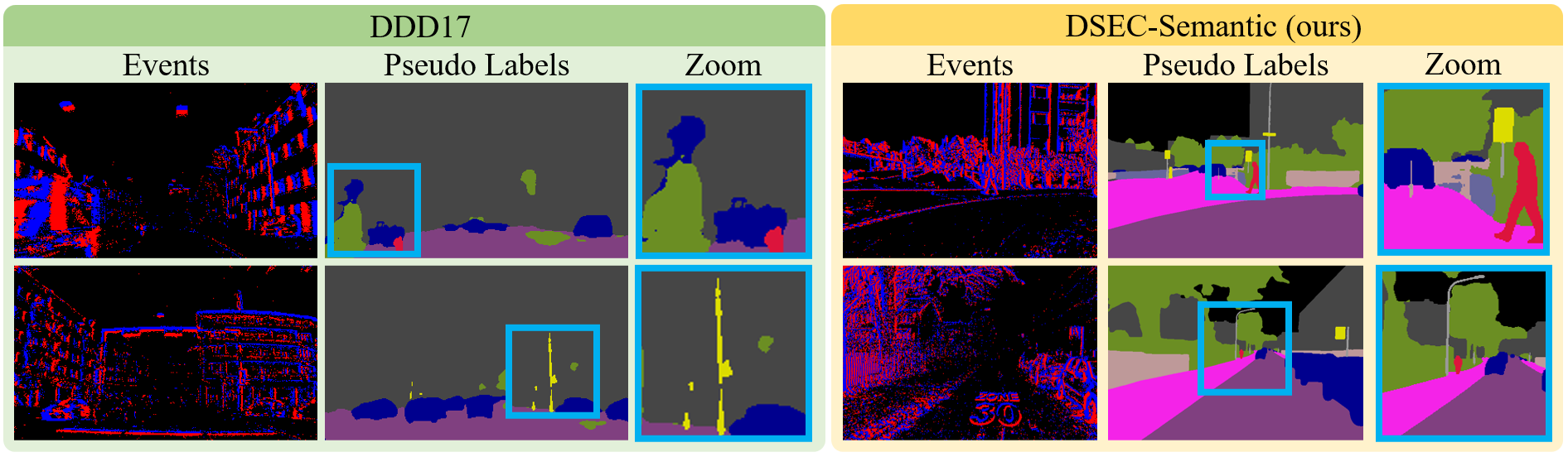}
\caption{
We release a new semantic segmentation dataset for the DSEC~\cite{Gehrig21ral} dataset. 
The pseudo labels are constructed based on the RGB images and a state-of-the-art frame-based segmentation network~\cite{Tao20arxiv}.
Compared to DDD17~\cite{Binas17icml,Alonso19cvprw} (left), our labels have a higher level of detail, seen in the zooms.
Additionally, our dataset includes more classes (11 classes) compared to~\cite{Alonso19cvprw} (6 classes).
}
\label{fig:dataset_overview}
\end{figure} 

%% file: floats/fig_dsec_uda_samples.tex
\begin{figure}[t!]
\centering
\includegraphics[width=0.99\textwidth]{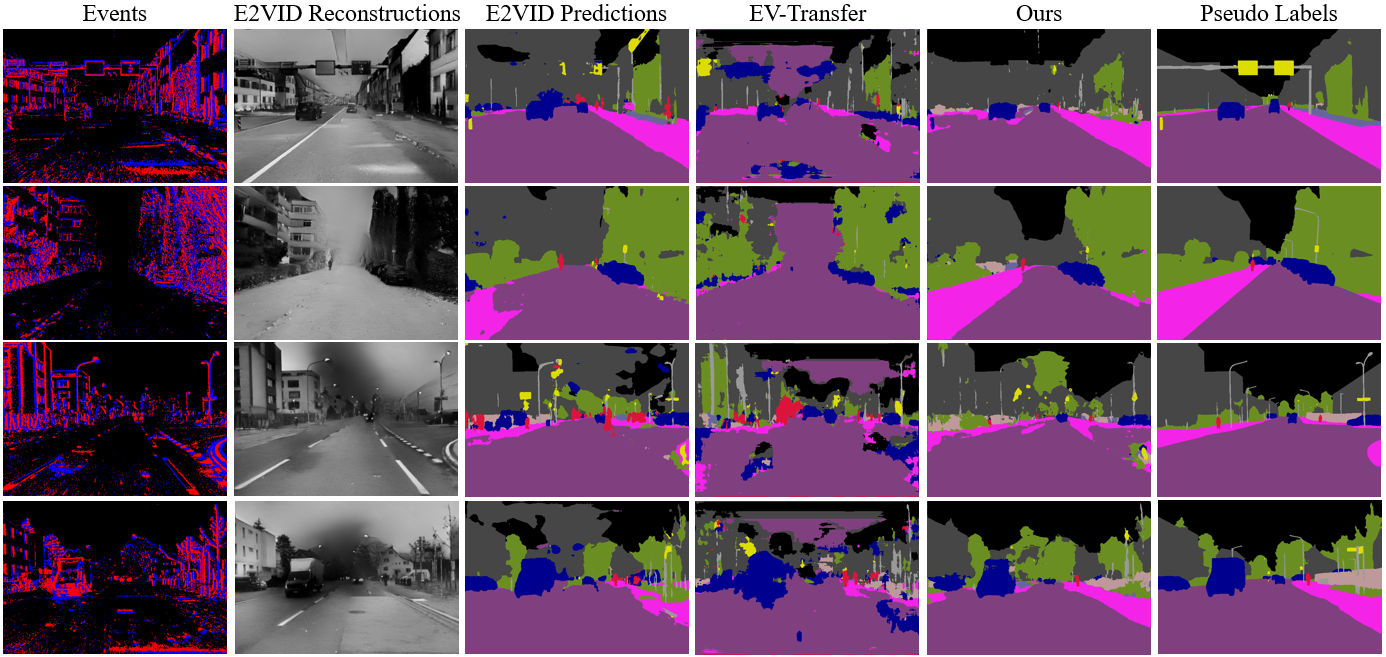}
\caption{
Qualitative samples on DSEC-Semantic for the UDA setting, i.e., no event labels are available during training.
Compared to EV-Transfer and E2VID, our method can more reliably predict smaller details such as persons.
}
\label{fig:dsec_uda_samples}
\end{figure} 

%% file: floats/table_dsec_uda.tex
\begin{table}[t!]
\caption{Performance of the UDA methods on DSEC-Semantic, which can leverage image labels and unpaired, unlabeled event data.
\todo{Results report the mean and standard deviation of 3 runs with different random seeds.}}
\centering
\scalebox{1}{
\begin{tabular}{m{2.7cm}C{2.7cm}C{2.7cm}>{\centering\arraybackslash}m{2.7cm}}
\hline
Method  & Labels & Accuracy [\%] $\uparrow$ & mIoU [\%] $\uparrow$ \\
 \hline
EV-Transfer~\cite{Messikommer22ral} & frames & \todo{60.50$\pm$2.50} & \todo{23.20$\pm$1.17} \\
E2VID~\cite{Rebecq19pami}           & frames & \todo{76.67$\pm$3.39} & \todo{40.70$\pm$3.38} \\
\textbf{ESS (ours)}                                & frames& \todo{\textbf{84.04$\pm$0.12}} & \todo{\textbf{44.87$\pm$0.51}}  \\
\hline
\end{tabular}
}
\label{tab:exp_dsec_uda}
\end{table}



%% file: floats/fig_dsec_supervised_samples.tex
\begin{figure}[t!]
\centering
\includegraphics[width=0.83\textwidth]{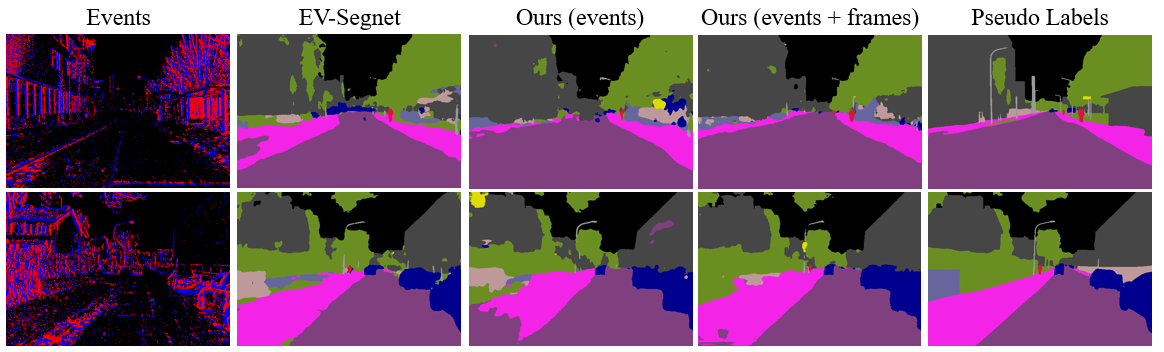}
\caption{
Qualitative samples on DSEC-Semantic in the supervised setting, i.e., event labels are available during training.
The combined training on image and event labels improves the semantic predictions.
Importantly, at test-time all methods only use events.
}
\label{fig:dsec_supervised_samples}
\end{figure} 

%% file: floats/table_dsec_supervised.tex
\begin{table}[t!]
\caption{
Results on DSEC-Semantic in the supervised setting, where event labels (events), image labels (labels), or both (events+frames) can be used.
}
\centering
\scalebox{1}{
\begin{tabular}{m{2.7cm}C{2.7cm}C{2.7cm}>{\centering\arraybackslash}m{2.7cm}}
\hline
Method  & Labels & Accuracy [\%] $\uparrow$ & mIOU [\%] $\uparrow$ \\
 \hline
EV-SegNet~\cite{Alonso19cvprw} & events        & 88.61 & 51.76 \\
\textbf{ESS (ours)}                           & frames        & 84.17 & 45.38  \\
\textbf{ESS (ours)}                           & events        & 89.25 & 51.57 \\
\textbf{ESS (ours)}                           & events+frames & \textbf{89.37} & \textbf{53.29}  \\
\hline
\end{tabular}
}
\label{tab:exp_dsec_supervised}
\end{table}

%% file: floats/table_combined_ablation.tex
\begin{table}
\parbox{.55\linewidth}{
\input{floats/table_dsec_ablation}
}
\hfill
\parbox{.4\linewidth}{
\input{floats/table_ddd17_alignment}
}
\end{table}

%% file: floats/table_dsec_ablation.tex
\caption{Ablation experiments on DSEC-Semantic in the UDA setting.}
\centering
\scalebox{1}{
\begin{tabular}{m{2.5cm}C{2.4cm}>{\centering\arraybackslash}m{1.7cm}}
\hline
Method  & Accuracy [\%]$\uparrow$ & mIoU [\%]$\uparrow$ \\
 \hline
w/o    $L_\text{cons. emb.}$    & 80.86 & 39.31 \\
w/o    $L_\text{cons. pred.}$   & 83.62 & 43.59 \\
w/o    $L_{\text{cons. task}}$  & 82.50 & 42.78 \\
w/o    skip connect.         & 78.79 & 38.08 \\
\textbf{ESS (ours)}                          & \todo{\textbf{84.04}} & \todo{\textbf{44.87}} \\
\hline
\end{tabular}
}
\label{tab:exp_dsec_ablation}

%% file: floats/table_ddd17_alignment.tex
\caption{
Alignment between image- and event-based predictions on DSEC-Semantic. Lower numbers mean better alignment.
}
\centering
\scalebox{1}{
\begin{tabular}{m{2.4cm}>{\centering\arraybackslash}m{2.2cm}}
\hline
Method  & Dissimilarity$\downarrow$ \\
 \hline
EV-Transfer~\cite{Messikommer22ral} & 0.120 \\
E2VID~\cite{Rebecq19pami}           & 0.073 \\
\textbf{ESS (ours)}                                & \textbf{0.025} \\
\hline
\end{tabular}
}
\label{tab:exp_ddd17_alignment}

%% file: sections/conclusion.tex
\section{Conclusion}
Event cameras promise to enhance the reliability of autonomous systems by improving the robustness of semantic segmentation networks in edge case scenarios such as during the night or at high speeds.
However, the lack of high-quality labeled datasets currently hinders the progress of event-based semantic segmentation.
In this work, we tackled this problem, by introducing ESS, which leverages large-scale, labeled image datasets for event-based semantic segmentation, without requiring event labels or paired events and images.
We thoroughly evaluated our method, both on the existing DDD17 benchmark, and the newly generated DSEC-Semantic benchmark, where we outperform existing state-of-the-art methods in UDA and supervised settings.
DSEC-Semantic is a large-scale event-based dataset for semantic segmentation, with high-quality, fine-grained semantic labels, which will spur further research in event-based semantic scene understanding.
While only evaluated for semantic segmentation, we believe that these performance gains can be transferred to other tasks. Our method unlocks the virtually unlimited supply of image-based datasets for event-based vision, enabling the exploration of previously inaccessible research fields for event cameras, such as panoptic segmentation, video captioning, action recognition etc.\\

%% file: sections/acknowledgment.tex
\par
\textbf{Acknowledgment} This work was supported by the National Centre of Competence in Research
(NCCR) Robotics through the Swiss National Science Foundation (SNSF)
and the European Research Council (ERC) under grant agreement No. 864042 (AGILEFLIGHT).

%% file: sections/supplementary.tex
\section*{\Large \bf Supplementary: ESS: Learning Event-based Semantic Segmentation from Still Images}
\section{DSEC-Semantic}
Our newly introduced event-based semantic segmentatation dataset, termed ~\textit{DSEC-Semantic}, is constructed based on sequences of the large-scale DSEC~\cite{Gehrig21ral} dataset, see Fig.~\ref{fig:dsec_overview}.
\todo{To generate the semantic labels, we first warp the images from the left frame-based camera  with a resolution of 1440 $\times$ 1080 to the view of the left event camera with a resolution of 640 $\times$ 480. 
The last 40 rows are then cropped since the frame-based camera does not capture these regions.
Thus, the the DSEC-semantic labels have a resolution of 640 $\times$ 440.
In a second step, we apply a state-of-the-art semantic segmentation method~\cite{Tao20arxiv} to the warped images to generate the labels.
We use pre-trained weights provided by the author.}

\todo{By doing so, we obtain fine-grained labels for 19 classes in the first place, which have the same classes than the Cityscapes labels for evaluation. 
We then further convert the 19 class labels into 11 classes (background, building, fence, person, pole, road, sidewalk, vegetation, car, wall, and traffic sign) for our experiments.
Since frame cameras suffer from image degradation in challenging illumination scenes, we only label a subset of sequences of the DSEC dataset which are recorded during the day to ensure high-quality labels.}
For the training set, we labeled 8082 frames of the following sequences: 'zurich\_city\_00\_a', 'zurich\_city\_01\_a', `zurich\_city\_02\_a', `zurich\_city\_04\_a', `zurich\_city\_05\_a', `zurich\_city\_06\_a', `zurich\_city\_07\_a', `zurich\_city\_08\_a'.
For the test set, we generated labels for 2809 frames of the following sequences: `zurich\_city\_13\_a', `zurich\_city\_14\_c', `zurich\_city\_15\_a'.\\
\todo{The dataset and detailed instructions are available at \url{https://dsec.ifi.uzh.ch/dsec-semantic/}}

\input{floats/fig_dsec_overview}

\section{Event Representation}
We convert an event stream $\mathcal{E}$ to a sequence of grid-like representations~\cite{Gehrig19iccv}, such as \emph{voxel grids}~\cite{Zhu19cvpr} $\mathbf{V}_k$. 
Each voxel grid is constructed from non-overlapping windows $\mathcal{E}_k$ each with a fixed number of events
\begin{equation}
    \mathbf{V}_k(x,y,t) = \sum_{e_j\in \mathcal{E}_k} p_j\delta(x_j-x)\delta(y_j-y)\max\{1-\vert t^*_j-t\vert, 0\},
\end{equation}
where $\delta$ is the Kronecker delta and $t^*_j = (B-1)\frac{t_j - t_0}{\Delta T}$ where $B$ is the number of bins, $\Delta T$ is the time window of events and $t_0$ is the time of the first event in the window. 

\section{Network Architecture}
Our network is a fully convolutional network inspired by the U-Net~\cite{Ronneberger15icmicci} architecture. 
We use an E2VID encoder $E_\text{E2VID}$ and an E2VID decoder $D_\text{E2VID}$ as illustrated in Fig. 4 of~\cite{Rebecq19pami} with the pre-trained weights provided by the author. 
The E2VID encoder $E_\text{E2VID}$ includes a head layer $\mathcal{H}$ and three recurrent encoder layers $\mathcal{E}^i$ with  $(i=0,1,2)$. 
We use the outputs of these three encoder layers as the recurrent, multi-scale embedding $\mathbf{z}_\text{event}$. 
The E2VID decoder $D_\text{E2VID}$ consists of the remaining two residual blocks $\mathcal{R}^j$, three decoder layers $\mathcal{D}^l$, and the final images prediction layer $\mathcal{P}$. 
For the image encoder $E_\text{img}$, we use the first layers up to the sixth residual block of ResNet-18~\cite{He16cvpr} without the first max-pooling layer. 
We use the outputs of the second and fourth residual blocks as skip connections for the task network. 
The encoder weights are initialized with parameters from ImageNet~\cite{Russakovsky15ijcv}. 
The task network $T$ consists of five residual blocks followed by seven convolution layers, and three upsampling layers lie in between. 
We use concatenation for the skip connection and nearest-neighbor interpolation with an upsampling factor of two for each upsampling layer.

\section{Training Details}
\noindent\textbf{DDD17}
For the experiments on DDD17, we use Cityscapes~\cite{Cordts16cvpr} as the labeled source domain and DDD17 as the unlabeled target domain. 
For each sample, we convert the events into a sequence of 20 voxel grids, each with 32'000 events.
The hyper-parameters $\lambda _{1}$, $\lambda _{2}$, $\lambda _{3}$, and $\lambda _{4}$ are set as 1, 0.01, 1, and 0.01, respectively.
We set the learning rates as $1\times 10^{-5}$ for $E_\text{img}$ and $1\times 10^{-4}$ for $T$.
We empirically found that having a smaller learning rate on $E_\text{img}$ and activating the accumulation of gradients for $E_\text{img}$ in the first stage help improve the results.
We train our model using the RAdam optimizer~\cite{Liu18iclr} with a batch-size of 16 for 50'000 iterations.    
Additionally, for the comparison with E2VID~\cite{Rebecq19pami} in the UDA setting, we retrain the image encoder and task network (forming a U-Net) on grayscale images and labels from the Cityscapes dataset~\cite{Cordts16cvpr}. 
Similar to our method, we train \cite{Messikommer22ral} in our UDA setting with the same source and target domains.

\noindent\textbf{DSEC-Semantic}
Similar to the experiments on DDD17, we leverage the Cityscapes datasets as the labeled source dataset.
The difference is that we now use the DSEC-Semantic dataset as the target domain. 
We increase the number of events per voxel grid to 100'000 due to the higher resolution.
To ensure the capturing of enough events at the beginning, we remove the first six samples of each sequence.
For computational reasons, we further skip every second sample of a selected sequence, which results in a training set of size 4017 and a test set of size 1395.
The hyper-parameters $\lambda _{1}$, $\lambda _{2}$, $\lambda _{3}$, and $\lambda _{4}$ are now set as 1, 1, 1, and 1, respectively.
We use the same RAdam optimizer to train our model with a larger learning rate of $5\times 10^{-4}$ (for both $E_\text{img}$ and $T$), and a smaller batch-size of 8, for 25'000 iterations.

\section{E2VID Driving Dataset}
To show that our method also works with completely unpaired and unlabeled data, we have applied it to the E2VID dataset~\cite{Rebecq19pami}, which contains driving sequences.
The dataset features events recorded with a Samsung DVS Gen3, and images recorded with a Huawei P20. 
Both cameras were mounted behind a car windshield, however, neither a external calibration nor a time synchronization is available.
Thus this dataset contains completely unpaired and unlabeled events.
Nevertheless, our method can learn the task on the image of Cityscapes and transfer it to the E2VID dataset, as shown in Fig.~\ref{fig:e2vid_uda_samples}.
\input{floats/fig_e2vid_uda_samples}

%% file: floats/fig_dsec_overview.tex
\begin{figure}[t!]
\centering
\includegraphics[width=0.99\textwidth]{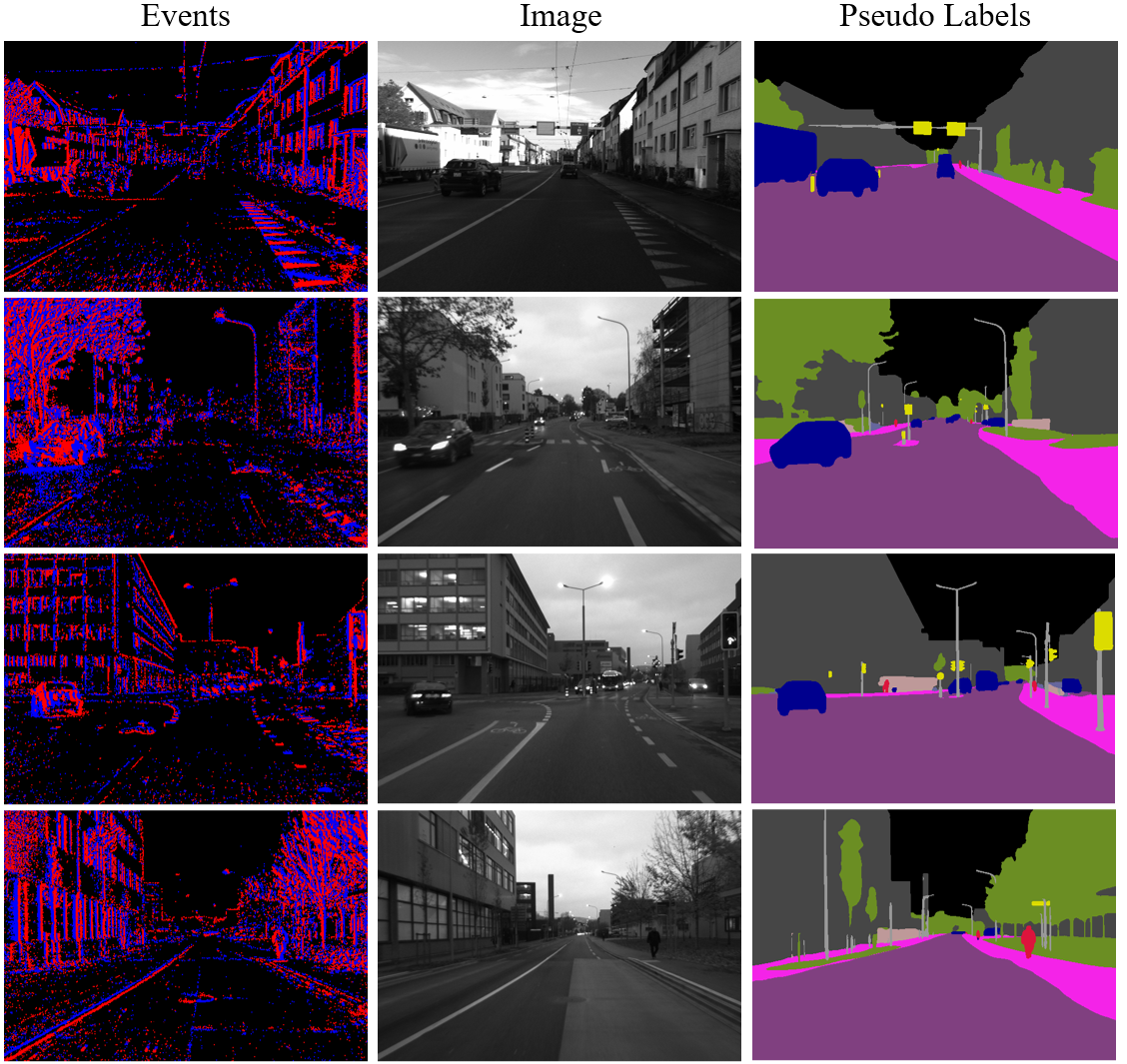}
\caption{
We release a new semantic segmentation dataset for the DSEC~\cite{Gehrig21ral} dataset containing accurate and fine-grained labels. 
The pseudo labels are constructed based on the RGB images and a state-of-the-art frame-based segmentation network~\cite{Tao20arxiv}.
}
\label{fig:dsec_overview}
\end{figure} 

%% file: floats/fig_e2vid_uda_samples.tex
\begin{figure}[t!]
\centering
\includegraphics[width=0.99\textwidth]{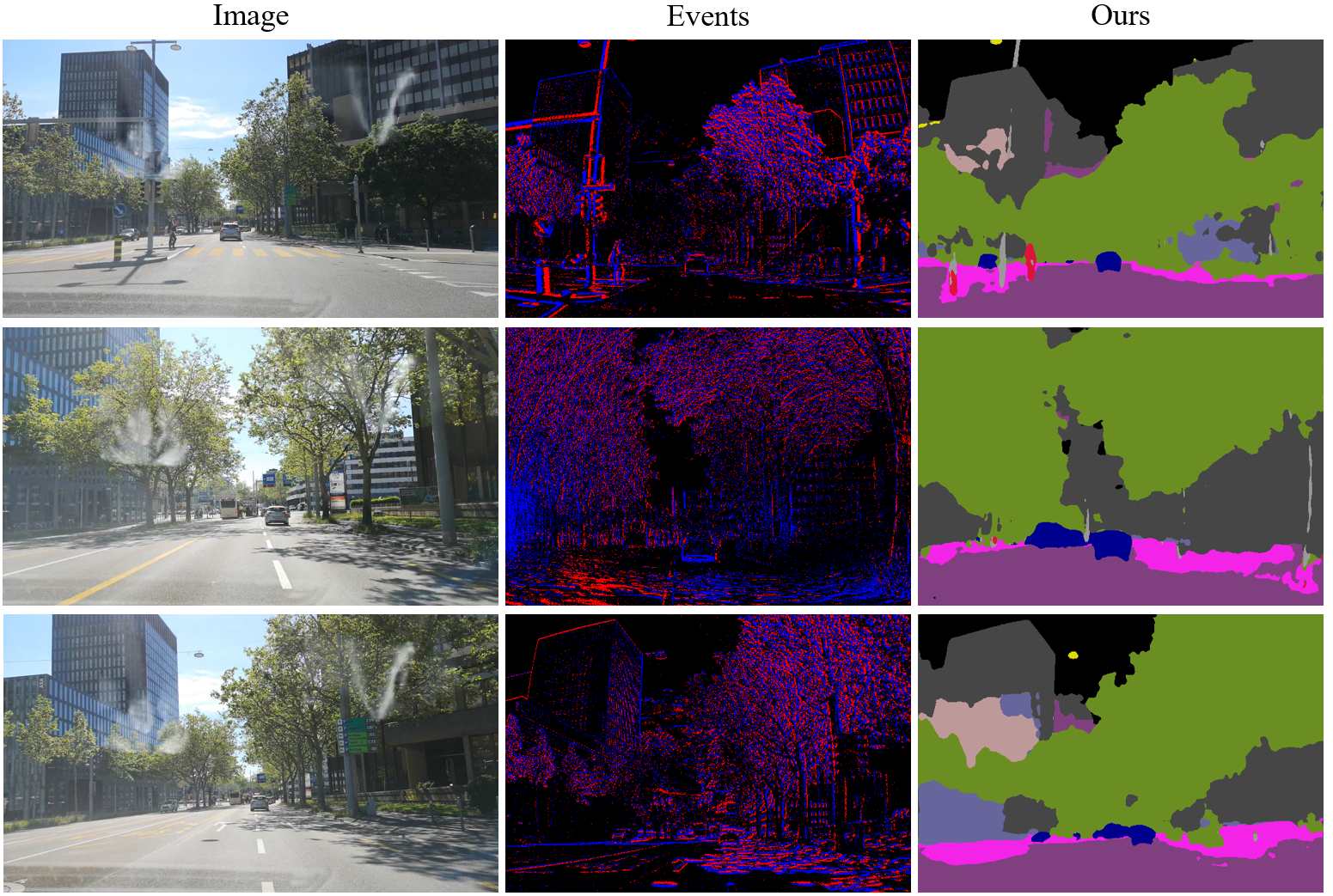}
\caption{
Qualitative samples on E2VID dataset for the UDA setting, i.e., no event labels are available during training.
There are no synchronized and aligned image and events available, thus we have selected the image in the dataset closest to the scene captured by the events.
}
\label{fig:e2vid_uda_samples}
\end{figure} 

%% file: main.bbl
\begin{thebibliography}{10}
\providecommand{\url}[1]{\texttt{#1}}
\providecommand{\urlprefix}{URL }
\providecommand{\doi}[1]{https://doi.org/#1}

\bibitem{Alonso19cvprw}
Alonso, I., Murillo, A.C.: {EV-SegNet}: Semantic segmentation for event-based
  cameras. In: {IEEE} Conf. Comput. Vis. Pattern Recog. Workshops (CVPRW)
  (2019)

\bibitem{Bardow16cvpr}
Bardow, P., Davison, A.J., Leutenegger, S.: Simultaneous optical flow and
  intensity estimation from an event camera. In: {IEEE} Conf. Comput. Vis.
  Pattern Recog. (CVPR). pp. 884--892 (2016). \doi{10.1109/CVPR.2016.102}

\bibitem{Binas17icml}
Binas, J., Neil, D., Liu, S.C., Delbruck, T.: {DDD}17: End-to-end {DAVIS}
  driving dataset. In: {ICML} Workshop on Machine Learning for Autonomous
  Vehicles (2017)

\bibitem{Brandli14ssc}
Brandli, C., Berner, R., Yang, M., Liu, S.C., Delbruck, T.: A 240x180 130{dB}
  3$\mu$s latency global shutter spatiotemporal vision sensor. {IEEE} J.
  Solid-State Circuits  \textbf{49}(10),  2333--2341 (2014).
  \doi{10.1109/JSSC.2014.2342715}

\bibitem{Chollet17cvpr}
Chollet, F.: Xception: Deep learning with depthwise separable convolutions. In:
  {IEEE} Conf. Comput. Vis. Pattern Recog. (CVPR). pp. 1800--1807 (2017).
  \doi{10.1109/CVPR.2017.195}

\bibitem{Cordts16cvpr}
Cordts, M., Omran, M., Ramos, S., Rehfeld, T., Enzweiler, M., Benenson, R.,
  Franke, U., Roth, S., Schiele, B.: The cityscapes dataset for semantic urban
  scene understanding. In: {IEEE} Conf. Comput. Vis. Pattern Recog. (CVPR)
  (2016)

\bibitem{Dosovitskiy17corl}
Dosovitskiy, A., Ros, G., Codevilla, F., Lopez, A., Koltun, V.: {CARLA}: {An}
  open urban driving simulator. In: Conf. on Robotics Learning (CoRL) (2017)

\bibitem{Falanga20Science}
Falanga, D., Kleber, K., Scaramuzza, D.: Dynamic obstacle avoidance for
  quadrotors with event cameras. Science Robotics  \textbf{5}(40),  eaaz9712
  (2020). \doi{10.1126/scirobotics.aaz9712}

\bibitem{Gallego20pami}
Gallego, G., Delbruck, T., Orchard, G., Bartolozzi, C., Taba, B., Censi, A.,
  Leutenegger, S., Davison, A., Conradt, J., Daniilidis, K., Scaramuzza, D.:
  Event-based vision: A survey. {IEEE} Trans. Pattern Anal. Mach. Intell.
  (2020). \doi{10.1109/TPAMI.2020.3008413}

\bibitem{Gehrig20cvpr}
Gehrig, D., Gehrig, M., Hidalgo-Carri\'o, J., Scaramuzza, D.: {V}ideo to
  {E}vents: Recycling video datasets for event cameras. In: {IEEE} Conf.
  Comput. Vis. Pattern Recog. (CVPR) (2020)

\bibitem{Gehrig19iccv}
Gehrig, D., Loquercio, A., Derpanis, K.G., Scaramuzza, D.: End-to-end learning
  of representations for asynchronous event-based data. In: Int. Conf. Comput.
  Vis. (ICCV) (2019)

\bibitem{RAL21Gehrig}
Gehrig, D., Rüegg, M., Gehrig, M., Hidalgo-Carrio, J., Scaramuzza, D.:
  Combining events and frames using recurrent asynchronous multimodal networks
  for monocular depth prediction. {IEEE} Robotic and Automation Letters. (RA-L)
   (2021)

\bibitem{Gehrig21ral}
Gehrig, M., Aarents, W., Gehrig, D., Scaramuzza, D.: Dsec: A stereo event
  camera dataset for driving scenarios. In: IEEE Robotics and Automation
  Letters (2021). \doi{10.1109/LRA.2021.3068942}

\bibitem{He16cvpr}
He, K., Zhang, X., Ren, S., Sun, J.: Deep residual learning for image
  recognition. In: {IEEE} Conf. Comput. Vis. Pattern Recog. (CVPR). pp.
  770--778 (2016). \doi{10.1109/cvpr.2016.90}

\bibitem{Yuhunang20eccv}
Hu, Y., Delbruck, T., Liu, S.C.: Learning to exploit multiple vision modalities
  by using grafted networks. In: Eur. Conf. Comput. Vis. (ECCV) (2020)

\bibitem{Hidalgo20threedv}
Javier Hidalgo-Carrio, D.G., Scaramuzza, D.: Learning monocular dense depth
  from events. {IEEE} International Conference on 3D Vision.(3DV)  (2020)

\bibitem{Liu18iclr}
Liu, L., Jiang, H., He, P., Chen, W., Liu, X., Gao, J., Han, J.: On the
  variance of the adaptive learning rate and beyond. In: Int. Conf. Learn.
  Representations ({ICLR}) (2020)

\bibitem{Maqueda18cvpr}
Maqueda, A.I., Loquercio, A., Gallego, G., Garc\'ia, N., Scaramuzza, D.:
  Event-based vision meets deep learning on steering prediction for
  self-driving cars. In: {IEEE} Conf. Comput. Vis. Pattern Recog. (CVPR). pp.
  5419--5427 (2018). \doi{10.1109/CVPR.2018.00568}

\bibitem{Messikommer22ral}
Messikommer, N., Gehrig, D., Gehrig, M., Scaramuzza, D.: Bridging the gap
  between events and frames through unsupervised domain adaptation. In: {IEEE}
  Robot. Autom. Lett. (2022)

\bibitem{muglikar213dv}
Muglikar, M., Moeys, D., Scaramuzza, D.: Event-guided depth sensing. In: {IEEE}
  International Conference on 3D Vision.(3DV) (December 2021)

\bibitem{Perot20nips}
Perot, E., de~Tournemire, P., Nitti, D., Masci, J., Sironi, A.: Learning to
  detect objects with a 1 megapixel event camera. In: Conf. Neural Inf.
  Process. Syst. (NIPS) (2020)

\bibitem{Rebecq19pami}
Rebecq, H., Ranftl, R., Koltun, V., Scaramuzza, D.: High speed and high dynamic
  range video with an event camera. {IEEE} Trans. Pattern Anal. Mach. Intell.
  (2019). \doi{10.1109/TPAMI.2019.2963386}

\bibitem{Reinbacher16bmvc}
Reinbacher, C., Graber, G., Pock, T.: Real-time intensity-image reconstruction
  for event cameras using manifold regularisation. In: British Mach. Vis. Conf.
  (BMVC) (2016). \doi{10.5244/C.30.9}

\bibitem{Ronneberger15icmicci}
Ronneberger, O., Fischer, P., Brox, T.: {U}-net: Convolutional networks for
  biomedical image segmentation. In: International Conference on Medical Image
  Computing and Computer-Assisted Intervention (2015)

\bibitem{Rosinol18ral}
{Rosinol Vidal}, A., Rebecq, H., Horstschaefer, T., Scaramuzza, D.: Ultimate
  {SLAM}? combining events, images, and {IMU} for robust visual {SLAM} in {HDR}
  and high speed scenarios. {IEEE} Robot. Autom. Lett.  \textbf{3}(2),
  994--1001 (Apr 2018). \doi{10.1109/LRA.2018.2793357}

\bibitem{Russakovsky15ijcv}
Russakovsky, O., Deng, J., Su, H., Krause, J., Satheesh, S., Ma, S., Huang, Z.,
  Karpathy, A., Khosla, A., Bernstein, M., Berg, A.C., Li, F.F.: {ImageNet}
  large scale visual recognition challenge. Int. J. Comput. Vis.
  \textbf{115}(3),  211--252 (Apr 2015). \doi{10.1007/s11263-015-0816-y}

\bibitem{Tao20arxiv}
Tao, A., Sapra, K., Catanzaro, B.: Hierarchical multi-scale attention for
  semantic segmentation. In: ArXiv (2020)

\bibitem{Tulyakov21cvpr}
Tulyakov, S., Gehrig, D., Georgoulis, S., Erbach, J., Gehrig, M., Li, Y.,
  Scaramuzza, D.: {TimeLens}: Event-based video frame interpolation. {IEEE}
  Conf. Comput. Vis. Pattern Recog. (CVPR)  (2021)

\bibitem{Wang20cvpr}
Wang, L., Kim, T.K., Yoon, K.J.: Eventsr: From asynchronous events to image
  reconstruction, restoration, and super-resolution via end-to-end adversarial
  learning. In: {IEEE} Conf. Comput. Vis. Pattern Recog. (CVPR). pp. 8312--8322
  (2020)

\bibitem{Wang21iccv}
Wang, L., Chae, Y., Yoon, K.J.: Dual transfer learning for event-based end-task
  prediction via pluggable event to image translation. In: Int. Conf. Comput.
  Vis. (ICCV). pp. 2135--2145 (2021)

\bibitem{Wang21cvpr}
Wang, L., Chae, Y., Yoon, S.H., Kim, T.K., Yoon, K.J.: Evdistill: Asynchronous
  events to end-task learning via bidirectional reconstruction-guided
  cross-modal knowledge distillation. In: {IEEE} Conf. Comput. Vis. Pattern
  Recog. (CVPR) (2021)

\bibitem{Zhu17cvpr}
Zhu, A.Z., Atanasov, N., Daniilidis, K.: Event-based visual inertial odometry.
  In: {IEEE} Conf. Comput. Vis. Pattern Recog. (CVPR). pp. 5816--5824 (2017).
  \doi{10.1109/CVPR.2017.616}

\bibitem{Zhu19cvpr}
Zhu, A.Z., Yuan, L., Chaney, K., Daniilidis, K.: Unsupervised event-based
  learning of optical flow, depth, and egomotion. In: {IEEE} Conf. Comput. Vis.
  Pattern Recog. (CVPR) (2019)

\end{thebibliography}
